\pdfoutput=1
\documentclass[11pt]{article}

\usepackage[]{acl}

\usepackage{times}
\usepackage{latexsym}

\usepackage[T1]{fontenc}
\usepackage[utf8]{inputenc}

\usepackage{microtype}

\usepackage{graphicx}

\usepackage{caption}
\usepackage{subcaption}
\usepackage{amsmath,amssymb}
\usepackage{mathtools}
\usepackage[thinc]{esdiff}

\usepackage{amsfonts}

\usepackage{amsmath}

\usepackage{physics}

\usepackage{multirow}

\usepackage{booktabs}

\DeclareMathOperator*{\argmax}{arg\,max}

\usepackage{array}
\usepackage{lipsum}

\newcommand{\ntasks}{18 }
\newcommand{\medianlangs}{11 }

\newcommand{\maxlangs}{100 }

\title{Beyond Static Models and Test Sets: Benchmarking the Potential of Pre-trained Models Across Tasks and Languages}

\author{Kabir Ahuja\textsuperscript{1} \quad Sandipan Dandapat\textsuperscript{2} \quad  Sunayana Sitaram\textsuperscript{1}  \quad Monojit Choudhury\textsuperscript{2}\\
\textsuperscript{1} Microsoft Research, India \\
\textsuperscript{2} Microsoft R\&D, India \\
{\tt \small \{t-kabirahuja,sadandap,sunayana.sitaram,monojitc\}@microsoft.com}
}

\begin{document}
\maketitle
\begin{abstract}
Although recent Massively Multilingual Language Models (MMLMs) like mBERT and XLMR support around 100 languages, most existing multilingual NLP benchmarks provide evaluation data in only a handful of these languages with little linguistic diversity. We argue that this makes the existing practices in multilingual evaluation unreliable and does not provide a full picture of the performance of MMLMs across the linguistic landscape. We propose that the recent work done in Performance Prediction for NLP tasks can serve as a potential solution in fixing benchmarking in Multilingual NLP by utilizing features related to data and language typology to estimate the performance of an MMLM on different languages. We compare performance prediction with translating test data with a case study on four different multilingual datasets, and observe that these methods can provide reliable estimates of the performance that are often on-par with the translation based approaches, without the need for any additional translation as well as evaluation costs.

% Our results indicate that the performance prediction methods can often provide reliable estimates of the performance on different languages that are often on par with translation-based approaches, but does not require any additional costs for translations. %Our results indicate that performance prediction is comparable to the translation based approach without incurring any additional costs. 
\end{abstract}

\section{Introduction}

% 1. Intro : Describe the problem with the multilingual evaluation in a higher level namely lack of an appropriate number of languages, low linguistic diversity amongst those which are present. Give a higher level overview of the performance prediction as a potential solution and cite the relevant papers the two papers from Graham Neubig's group, Anirudh's paper and AAAI submission.
\begin{figure*}
    \centering
    \begin{subfigure}[t]{0.45\textwidth}
    \includegraphics[width=0.95\textwidth]{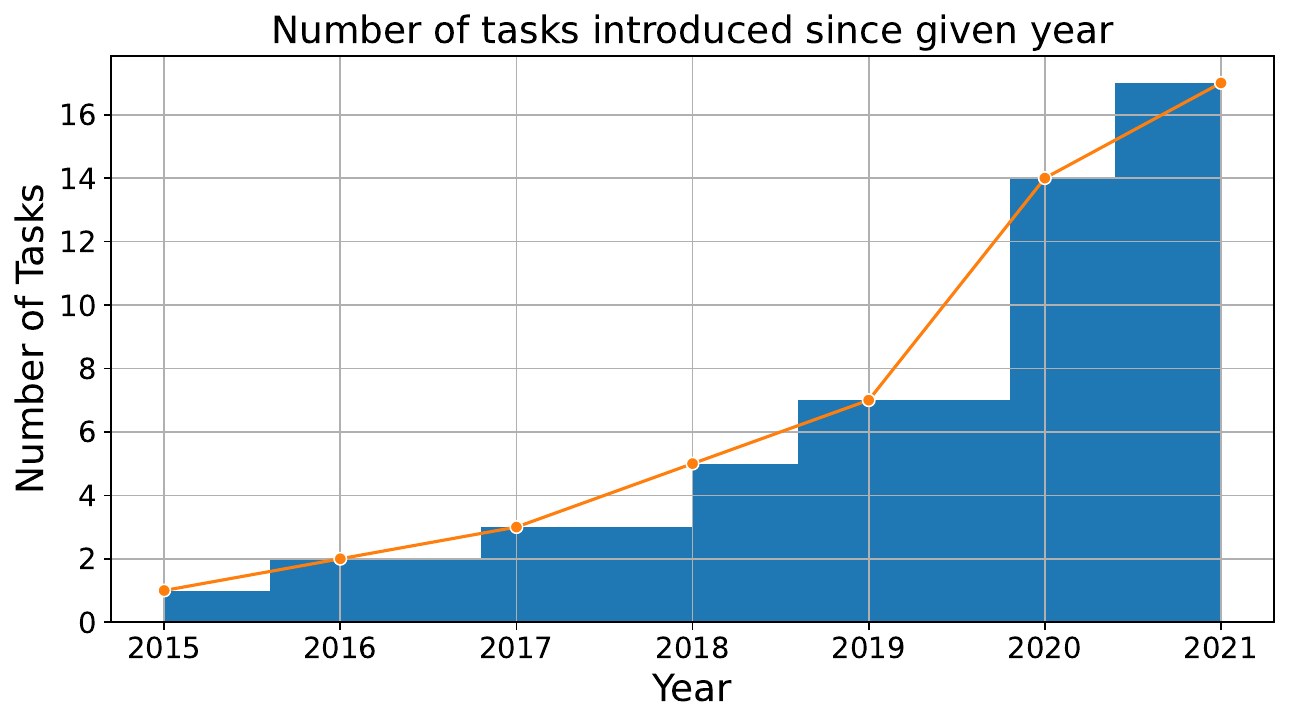}
    \caption{Cumulative distribution of the multilingual tasks proposed each year from 2015 to 2021.}
    \label{fig:yearwise_tasks}
    \end{subfigure}%
    \begin{subfigure}[t]{0.49\textwidth}
    \includegraphics[width=0.99\textwidth]{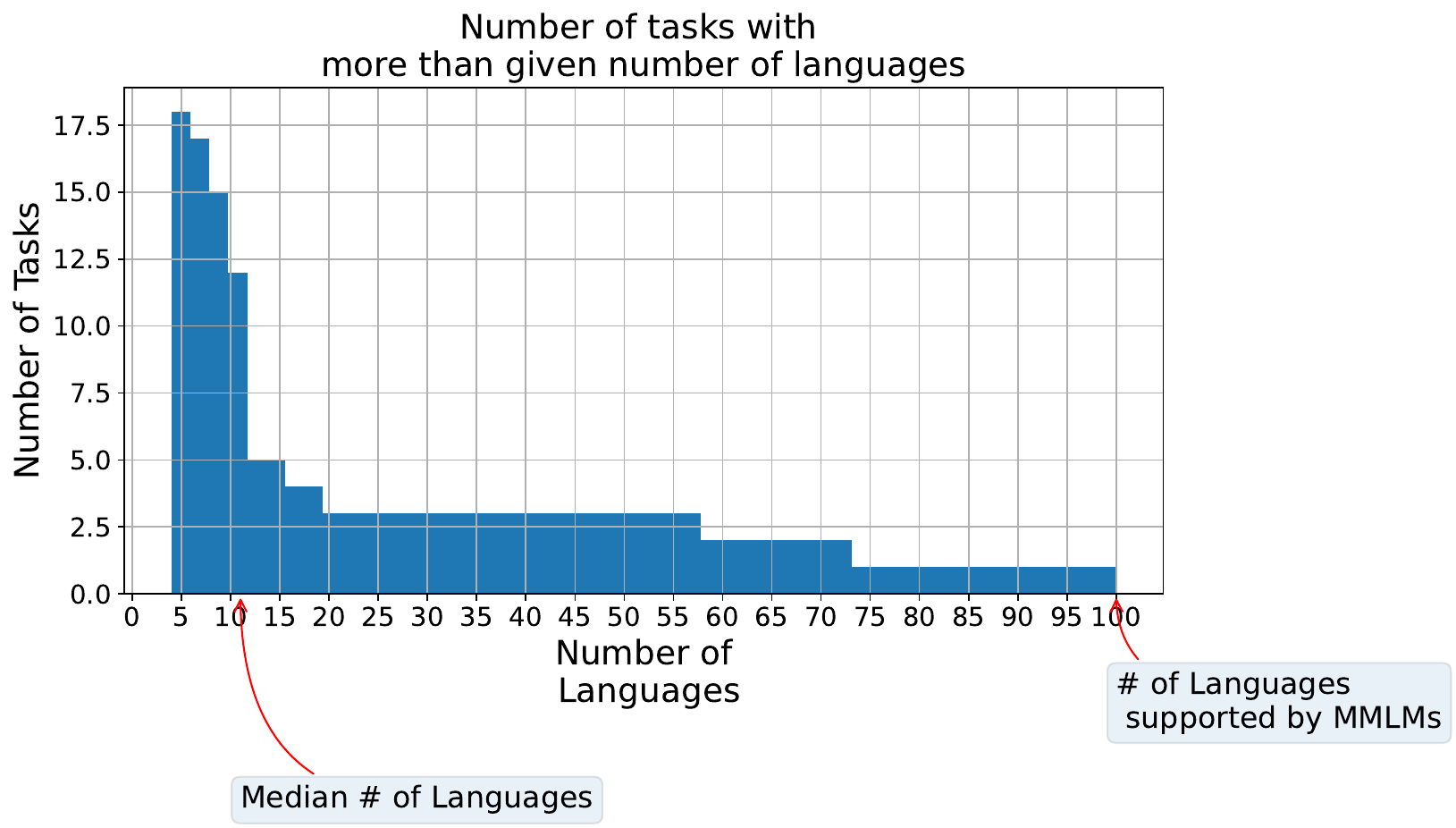}
    \caption{Reverse cumulative distribution for the number of languages available in different tasks.}
    \label{fig:tasks_langs}
    \end{subfigure}
    \begin{subfigure}[t]{0.98\textwidth}
    \includegraphics[width=0.98\textwidth]{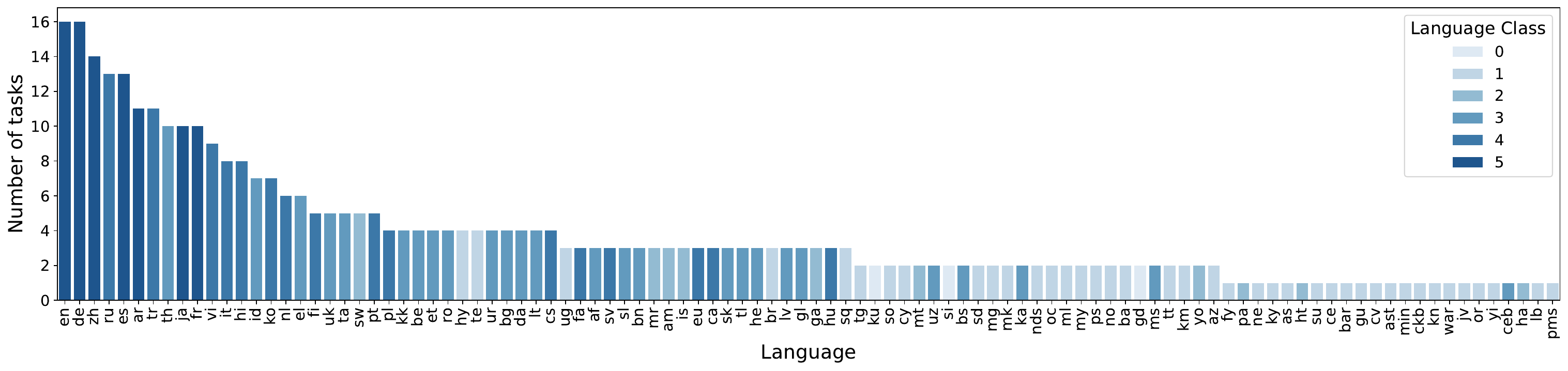}
    \caption{Number of multilingual tasks containing test data for each of the 106 languages supported by the MMLMs (mBERT, XLMR). The bars are shaded according to the class taxonomy proposed by \citet{joshi-etal-2020-state}.}
    \label{fig:lang_num_tasks}
    \end{subfigure}
\caption{}
\label{fig:task_lang_stats}
\end{figure*}
Recent years have seen a surge of transformer~\cite{vaswanietal2017} based Massively Multilingual Language Models (MMLMs) like mBERT \cite{devlin-etal-2019-bert} , XLM-RoBERTa (XLMR) \cite{conneau-etal-2020-unsupervised}, mT5 \cite{xue-etal-2021-mt5}, RemBERT \cite{chung2021rethinking}. These models are pretrained on varying amounts of data of around 100 linguistically diverse languages, and can in principle support fine-tuning on different NLP tasks for these languages. %MMLMs have been shown to be surprisingly effective for zero-shot transfer i.e. once fine-tuned on a \textit{pivot} language, they tend to perform well on unseen \textit{target} languages \cite{pires-etal-2019-multilingual, wu-dredze-2019-beto}.

These MMLMs are primarily evaluated for their performance on Sequence Labelling \cite{nivre-etal-2020-universal, Pan2017},  Classification \cite{Conneau2018xnli, Yang2019paws-x, ponti-etal-2020-xcopa}, Question Answering \cite{artetxe2020cross, Lewis2020mlqa, Clark2020tydiqa} and Retrieval \cite{Artetxe2019massively, roy-etal-2020-lareqa, botha-etal-2020-entity} tasks. However, most these tasks often cover only a handful of the languages supported by the MMLMs, with most tasks having test sets in fewer than 20 languages (cf. Figure \ref{fig:tasks_langs}). 

%The only task that has representation of almost all the 100 languages is Structure Prediction task WikiANN \cite{Pan2017} which was made possible through the crowd-sourcing efforts for about 35 million articles and numerous engineering efforts for entity linking and refining the annotations. The only other tasks that contain contain test sets for more than 50 of the supported languages are Universal Dependencies benchmark \cite{nivre-etal-2020-universal} another structure prediction task made possible due to the presence of 200 treebanks produced by about 300 contributors, and the retrieval task Tatoeba \cite{Artetxe2019massively} created from the presence of translation pairs in about 92 languages. However, all the remaining tasks that we surveyed, especially the NLU tasks had evaluation datasets available for fewer than 20 languages. 

Evaluating on such benchmarks henceforth fails to provide a comprehensive picture of the model's performance across the linguistic landscape, as the performance of MMLMs has been shown to vary significantly with the amount of pre-training data available for a language \cite{wu-dredze-2020-languages}, as well according to the typological relatedness between the \textit{pivot} and \textit{target} languages \cite{lauscher-etal-2020-zero}.  While designing  benchmarks to contain test data for all 100 languages supported by the MMLMs is be the ideal standard for multilingual evaluation, doing so requires prohibitively large amount of human effort, time and money. %The only tasks which have been able to support a good fraction of these 100 langauges are the Sequence Labelling tasks WikiANN \cite{Pan2017} and Universal Dependencies\cite{nivre-etal-2020-universal} which were a result of huge engineering, crowd sourcing and domain expertise efforts, and the Tatoeba dataset created from the parallel translation database maintained since more than 10 years and consisted of contributions from tens of thousands of members. The nature and state of the most other NLU tasks makes it much harder to create such language exhaustive datasets.

Machine Translation can be one way to extend test sets in different benchmarks to a much larger set of languages. \citet{hu2020xtreme} provides pseudo test sets for tasks like XQUAD and XNLI, obtained by translating English test data into different languages, and shows reasonable estimates of the actual performance by evaluating on translated data but cautions about their reliability when the model is trained on translated data. The accuracy of translation based evaluation can be affected by the quality of translation and the technique incurs non-zero costs to obtain reliable translations. Moreover, transferring labels with translation might also be non-trivial for certain tasks like Part of Speech Tagging and Named Entity Recognition.

Recently, there has been some interest in predicting performance of NLP models without actually evaluating them on a test set. \citet{xia-etal-2020-predicting} showed that it is possible to build regression models that can accurately predict evaluation scores of NLP models under different experimental settings using various linguistic and dataset specific features. \citet{srinivasan2021predicting} showed promising results specifically for MMLMs towards predicting their performance on downstream tasks for different languages in zero-shot and few-shot settings, and \citet{ye-etal-2021-towards} propose methods for more reliable performance prediction by estimating confidence intervals as well as predicting fine-grained performance measures.

In this paper we argue that the performance prediction can be a possible avenue to address the current issues with Multilingual benchmarking by aiding in the estimation of performance of the MMLMs for the languages which lack any evaluation data for a given task. Not only this can help us give a better idea about the performance of a multilingual model on a task across a much larger set of languages and hence aiding in better model selection, but also enables applications in devising data collection strategies to maximize performance \cite{srinivasan2022litmus} as well as in selecting the representative set of languages for a benchmark \cite{xia-etal-2020-predicting}.

We present a case study demonstrating the effectiveness of performance prediction on four multilingual tasks, PAWS-X \cite{Yang2019paws-x} XNLI \cite{Conneau2018xnli}, XQUAD \cite{artetxe2020cross} and TyDiQA-GoldP \cite{Clark2020tydiqa} and show that it can often provide reliable estimates of the performance on different languages on par with evaluating them on translated test sets without any additional translation costs. We also demonstrate an additional use case of this method in selecting the best pivot language for fine-tuning the MMLM in order to maximize performance on some target language. To encourage research in this area and provide easy access for the community to utilize this framework, we will release our code and the datasets that we use for the case study.

% 2. Background / Problem: Go into detail about the problem, how the current multilingual evaluation is broken. Refer to the different statistics pertaining to the language coverage of various multilingual benchmarks, linguistic diversity and resource diversity. Further provide some examples of how this limited view of the linguistic landscape can obscure our point of view about the underrepresented languages. AmericasNLI https://arxiv.org/abs/2104.08726 is one such example where the performance on southern/central american languages on the NLI task is observed to be significantly worse for XLMR. Fairness paper by Monojit. We might also want to talk about tasks like TyDiQA and XCOPA which are linguistically diverse and how that alone might not be sufficient to get a full picture of the task. Cite examples where languages in same family have very different performance

\section{The Problem with Multilingual Benchmarking}
\label{sec:problem}
The rise in popularity of MMLMs like mBERT and XLMR have also lead to an increasing interest in creating different multilingual benchmarks to evaluate these models. We analyzed \ntasks different multilingual datasets proposed between the years 2015 to 2021, by searching and filtering for datasets containing the term \textit{Cross Lingual} in the Papers with Code Datasets repository.\footnote{https://paperswithcode.com/datasets} The types and language specific statistics of these studied benchmarks can be found in Table \ref{tab:all_tasks} in appendix.

As can be seen in Figure \ref{fig:yearwise_tasks}, there does appear to be an increasing trend in the number of multilingual datasets proposed each year, especially with a sharp increase observed during the year 2020. However, if we look at the number of languages covered by these different benchmarks (Figure \ref{fig:tasks_langs}), we see that most of the tasks have fewer than 20 languages supported with a median of \medianlangs languages per task which is substantially lower than the \maxlangs supported by the commonly used MMLMs.

% The only task that has representation of almost all the 100 languages is Sequence Labelling task WikiANN \cite{Pan2017} which was made possible through the crowd-sourcing efforts for about 35 million articles and numerous engineering efforts for entity linking and refining the annotations. The other two tasks that contain test sets for more than 50 of the supported languages are Universal Dependencies benchmark \cite{nivre-etal-2020-universal} another sequence labelling task made possible due to the presence of 200 treebanks produced by the domain expertise of about 300 contributors, and the retrieval task Tatoeba \cite{Artetxe2019massively} created from the presence of translation pairs in about 92 languages from the database maintained by the community of tens of thousands of contributors. However, all the remaining tasks that we surveyed, especially the NLU tasks we observed a dearth of supported languages.

The only tasks which have been able to support a large fraction of these 100 languages are the Sequence Labelling tasks WikiANN \cite{Pan2017} and Universal Dependencies\cite{nivre-etal-2020-universal} which were a result of huge engineering, crowd sourcing and domain expertise efforts, and the Tatoeba dataset created from the parallel translation database maintained since more than 10 years, consisting of contributions from tens of thousands of members. However, we observed a dearth of supported languages in the remaining tasks that we surveyed, especially in NLU tasks.

We also observe a clear lack of diversity in the selected languages across different multilingual datasets. Figure \ref{fig:lang_num_tasks} shows the number of tasks each language supported by the mBERT is present in and we observe a clear bias towards high resource languages, mostly covering class 4 and class 5 languages identified according to the taxonomy provided by \citet{joshi-etal-2020-state}. The low resource languages given by class 2 or lower are severely under-represented in the benchmarks where the most popular (in terms of number of tasks it appears in) class 2 language i.e. Swahili appears only in 5 out of \ntasks benchmarks.

\begin{figure}
    \centering
    \includegraphics[width=0.4\textwidth]{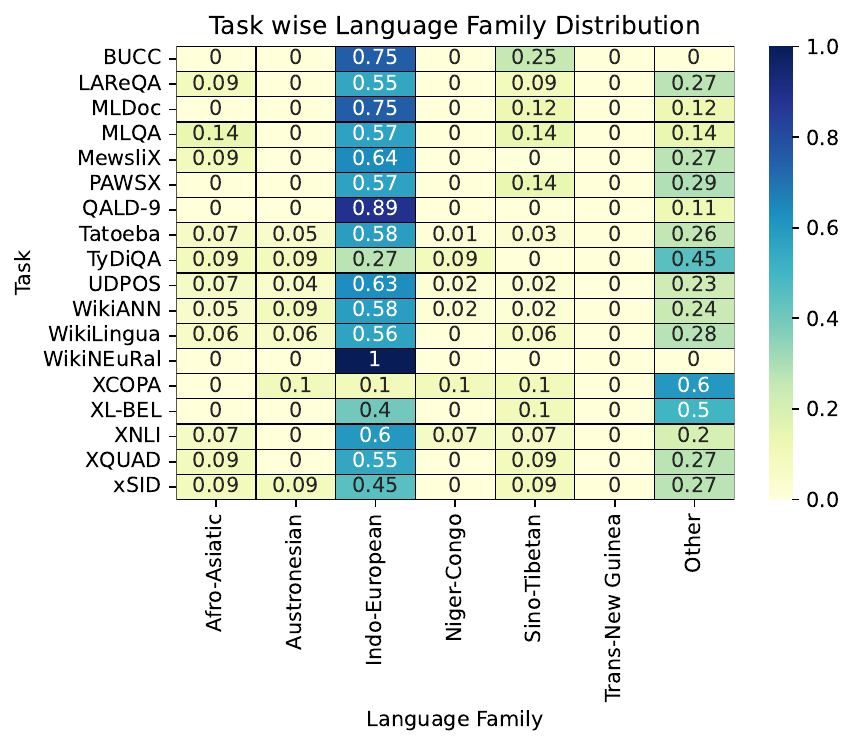}
    \caption{Task wise distribution of language families i.e. fraction of languages belonging to a particular language for a task.}% We consider the the major language families at the top level genetic groups, each of which have at least 5\% of the world languages according to Ethnologue.}
    \label{fig:lang_fam_dist}
\end{figure}

We also categorized the the languages into the 6 major language families at the top level genetic groups \footnote{https://www.ethnologue.com/guides/largest-families} each of which cover at least 5\% of the world's languages and plot language family wise representation of each task in Figure \ref{fig:lang_fam_dist}. Except a couple of benchmarks, the majority of the languages present in these tasks are Indo-European, with very little representation from all the other language families which have either comparable or a higher language coverage as Indo-European.

There have been some recent benchmarks that address this issue of language diversity. The TyDiQA \cite{Clark2020tydiqa} benchmark contains training and test datasets in 11 typologically diverse languages, covering 9 different language families. The XCOPA \cite{ponti-etal-2020-xcopa} benchmark for causal commonsense reasoning also selects a set of 10 languages with high genealogical and areal diversities.

While this is a step in the right direction and does give a much better idea about the performance of MMLMs over a diverse linguistic landscape, it is still difficult to cover through 10 or 11 languages all the factors that influence the performance of an MMLM like pre-training size \cite{wu-dredze-2020-languages, lauscher-etal-2020-zero}, typological relatedness (syntactic, genealogical, areal, phonological etc) between the source and pivot languages \cite{lauscher-etal-2020-zero, pires-etal-2019-multilingual}, sub-word overlap \cite{wu-dredze-2019-beto}, tokenizer quality  \cite{rust-etal-2021-good} etc. Through Performance Prediction as we will see in next section, we seek to estimate the performance of an MMLMs on different languages based on these factors. 

We would also like to point out that there are other problems with multilingual benchmarking as well. Recent multi-task multilingual benchmarks like X-GLUE \cite{liang-etal-2020-xglue}, XTREME \cite{hu2020xtreme} and XTREME-R \cite{ruder-etal-2021-xtreme} mainly provide training datasets for different tasks only in English and evaluate for zero-shot transfer to other languages. However, this standard of using English as a default pivot language was put in question by \citet{turc2021revisiting}, who showed empirically that German and Russian transfer more effectively to a set of diverse target languages. We shall see in the coming sections that the Performance Prediction approach can also be useful in identifying the best pivots for a target language.

% 3. Potential Solution : Introduce the task of performance prediction in NLP formally, cover zero-shot, few-shot, multi-pivot etc. cases in the general definition. Discuss various ways to define the features according to the previous work and potential approaches for training such models and evaluation methodology. Translating the test data can be considered as an alternative, but there are known problems with it which can be discussed.

\section{Performance Prediction for Multilingual Evaluation}

We define Performance Prediction as the task of predicting performance of a machine learning model on different configurations of training and test data. Consider a multilingual model $\mathcal{M}$ pre-trained on a set of languages $\mathcal{L}$, and a task $\mathfrak{T}$ containing training datasets $\mathcal{D}_{tr}^p$ in languages $p \in \mathcal{P}$ such that $\mathcal{P} \subset \mathcal{L}$ and test datasets $\mathcal{D}_{te}^t$ in languages  $t \in \mathcal{T}$ such that $\mathcal{T} \subset \mathcal{L}$. Following \citet{NIPS2009_f79921bb}, we assume that both $\mathcal{D}_{tr}^p$ and $\mathcal{D}_{te}^t$ are the subsets of a multi-view dataset $\mathcal{D}$ where each sample $(\vb{x}, \vb{y}) \in \mathcal{D}$ has multiple views (defined in terms of languages) of the same object i.e. $(\vb{x}, \vb{y}) \overset{\mathrm{def}}{=} \{(x^{l}, y^{l}) | \forall l \in \mathcal{L}$\} all of which are not observed. %From this assumption we imply that the datasets in different languages follow the same distribution.

A training configuration for fine-tuning $\mathcal{M}$ is given by the tuple $(\Pi, \Delta_{tr}^{\Pi})$, where $\Pi \subseteq \mathcal{P}$ and $\Delta_{tr}^{\Pi} = \bigcup\limits_{p \in \Pi}\mathcal{D}_{tr}^p$. The performance on the test set $\mathcal{D}_{te}^t$ for language $t \in \mathcal{T}$ when $\mathcal{M}$ is fine-tuned on $(\Pi, \Delta_{tr}^{\Pi})$ is denoted as $s_{\mathcal{M}, \mathfrak{T}, t, \mathcal{D}_{te}^t, \Pi, \Delta_{tr}^{\Pi}}$ or $s$ for clarity, given as:
\begin{equation}
    s = g(\mathcal{M}, \mathfrak{T}, t, \mathcal{D}_{te}^t, \Pi, \Delta_{tr}^{\Pi})
\end{equation}

In performance prediction we formulate estimating $g$ by a parametric function $f_{\theta}$ as a regression problem such that we can approximate $s$ for various configurations with reasonable accuracy, given by

\begin{equation}
    s \approx f_{\theta}([\phi(t); \phi(\Pi); \phi(\Pi, t); \phi(\Delta_{tr}^{\Pi})])
\end{equation}

% \begin{equation}
%     s \approx f_{\theta}([\phi(\mathcal{M}); \phi(t); \mathcal{D}_{te}^t; \phi(\Pi);\phi(\Pi, t); \phi(\Delta_{tr}^{\Pi})])
% \end{equation}
where $\phi(.)$ denotes the features representation of a given entity. Following \citet{xia-etal-2020-predicting}, we do not consider any features specific to $\mathcal{M}$ to focus more on how the performance varies for a given model with different data and language configurations.  Since the languages for which we are trying to predict the performance might not have any data (labelled or unlabelled available), we also skip features for $\mathcal{D}_{te}^t$ from the equation. Note, we do consider coupled features for training and test languages i.e. $\phi(\Pi, t)$ as the interaction between the two has been shown to be a strong indicator of the performance of such models \cite{lauscher-etal-2020-zero, wu-dredze-2019-beto}. 

% \begin{equation}
%     s \approx f_{\theta}([\phi(t); \phi(\Pi); \phi(\Pi, t); \phi(\Delta_{tr}^{\Pi})])
% \end{equation}

Different training setups for multilingual models can be seen as special cases of our formulation. For zero-shot transfer we set $\Pi = \{p\}$, such that $p \neq t$. This reduces the performance prediction problem to the one described in \citet{lauscher-etal-2020-zero}.
\begin{equation}
    s_{zs} \approx f_{\theta}([\phi(t); \phi(p); \phi(p, t); \phi(\Delta_{tr}^{\{p\}})])
\end{equation}

% Similarly the monolingual training setup as discussed in \citet{rust-etal-2021-good} can be obtained by setting $\Pi = \{t\}$, giving:
% \begin{equation}
%     s_{mono} \approx f_{\theta}([\phi(t); \phi(t); \phi(t, t); \phi(\Delta_{tr}^{\{t\}})])
% \end{equation}

There are many ways to represent the feature representations $\phi(.)$ that have been explored in previous work, including pre-training data size, typological relatedness between the pivot and target languages and more. For a complete list of features that we use in our experiments, refer to Table \ref{tab:features}.

\begin{table}[]
    \small
    \centering
    \begin{tabular}{m{1cm}m{2cm}m{2cm}}
        \toprule
        \textbf{Type} & \textbf{Features} & \textbf{Reference} \\
        \midrule
        \multirow{2}{*}{$\phi(t)$} & Pre-training Size of $t$ & \citet{srinivasan2021predicting, lauscher-etal-2020-zero}\\
        \cmidrule{2-3}
        & Tokenizer Quality for $t$ & \citet{rust-etal-2021-good}\\
        \midrule
        \multirow{1}{*}{$\phi(\Pi)$} & Pre-training size of every $p \in \Pi$  & \\
        \midrule
        \multirow{2}{*}{$\phi(\Pi, t)$} & Subword Overlap between $p$ and $t$ for $p \in \Pi$ & \citet{lin-etal-2019-choosing,xia-etal-2020-predicting, srinivasan2021predicting} \\
        \cmidrule{2-3}
        & Relatedness between lang2vec \cite{littell-etal-2017-uriel} features & \citet{lin-etal-2019-choosing,xia-etal-2020-predicting,lauscher-etal-2020-zero, srinivasan2021predicting}\\
        \midrule
        $\phi(\Delta_{tr}^{\Pi})$ & Training size $|\mathcal{D}_{tr}^p|$ of each language $p \in \Pi$ & \cite{lin-etal-2019-choosing,xia-etal-2020-predicting, srinivasan2021predicting}\\
        \bottomrule
    \end{tabular}
    \caption{Features used to represent the languages and datasets used. For more details refer to Section \ref{sec:feat_desc} in Appendix.}
    \label{tab:features}
\end{table}

% 4. Case Study : Consider two tasks a linguistically diverse one like TyDIQA and a lesser diverse task like XQUAD, define the features and show LOLO errors and predictions over a large set of languages. The quality of predictions can be evaluated by comparing the predicted performance with the performance on translated data on few of those languages.

\section{Case Study}

\begin{table*}[]
\centering
%\small
\begin{tabular}{p{2.5cm}p{1.5cm}p{1.5cm}p{1.5cm}p{1.5cm}}
\toprule
\textbf{Task} &  \textbf{Baseline} & \textbf{Translate} & \multicolumn{2}{c}{\textbf{Performance Predictors }} \\
\cline{4-5}
& & &  \textbf{XGBoost} &  \textbf{Group Lasso} \\
\midrule
PAWS-X &                7.18 &                      3.85 &     5.46 &         \textbf{3.06} \\
XNLI &                5.32 &                      \textbf{2.70} &     3.36 &         3.93 \\
XQUAD &                6.89 &                      \textbf{3.42} &     5.41 &         4.53 \\
TyDiQA-GoldP &                7.82 &                      7.77 &     5.04 &         \textbf{4.73} \\
\bottomrule
\end{tabular}
\caption{Mean Absolute Errors (MAE) (scaled by 100 for readability) on the the three tasks for different methods of estimating performance.}
\label{tab:main_results}
\end{table*}

To demonstrate the effectiveness of Performance Prediction in estimating the performance on different languages, we evaluate the approach on classification tasks i.e. PAWS-X and XNLI, and two Question Answering tasks XQUAD and TyDiQA-GoldP. We choose these tasks as their labels are transferable via translation, so we can compare our method with the automatic translation based approach. TyDiQA-GoldP has test sets for different languages created independently to combat the \textit{translationese} problem \cite{clark-etal-2020-tydi}, while the other three have English test sets manually translated to the other languages.% This also enables us to investigate the reliability of translation based approaches to our performance prediction approach.

\subsection{Experimental Setup}
For all the three tasks we try to estimate zero-shot performance of a fine-tuned mBERT model i.e. $s_{zs}$ on different languages. For PAWS-X, XNLI and XQUAD we have training data present only in English i.e. $\Pi = \{en\}$ always, but TyDiQA-GoldP contains training sets in 9 different languages and we predict transfer from all of those. To train Performance Prediction models we use the performance data for mBERT provided in \citet{hu2020xtreme} as well as train our own models when required and evaluate the performance on test dataset of different languages. The performance prediction models are evaluated using a leave one out strategy also called \textit{Leave One Language Out} (LOLO) as used in \citet{lauscher-etal-2020-zero,srinivasan2021predicting}, where we use the performance data of target languages in the set $\mathcal{T} - \{t\}$ to predict the performance on a language $t$ and do this for all $t \in \mathcal{T}$.

\subsection{Methods}
We compare the following methods for estimating the performance:

\noindent \textbf{1. Average Score Baseline}: In this method, to estimate the performance on a target language $t$ we simply take a mean of the model's performance on the remaining $\mathcal{T} - \{t\}$ languages. Although conceptually simple, this is an unbiased estimate for the expected performance of the MMLM on different languages.%, and hence it can be useful to compare if the other methods can provide more closer estimates than the expected value.

\noindent \textbf{2. Translate}: To estimate the performance on language $t$ with this method, we automatically translate the test data in one of the languages $t' \in \mathcal{T} - \{t\}$ ,\footnote{for our experiments we use $t' = p$ i.e. we use test data in pivot language which is often English to translate to $t$} to the target language $t$ and evaluate the fine-tuned MMLM on the translated data. The performance on this pseudo-test set is used as the estimate of the actual performance. We use the Azure Translator\footnote{https://azure.microsoft.com/en-us/services/cognitive-services/translator/} to translate the test sets.

\noindent \textbf{3. Performance Predictors}: We consider  two different regression models to estimate the performance in our experiments.

i) \textbf{XGBoost}: We use the popular Tree Boosting algorithm XGBoost for solving the regression problem, which has been previously shown to achieve impressive results on the task \cite{xia-etal-2020-predicting, srinivasan2021predicting}.

ii) \textbf{Group Lasso}: Group Lasso \cite{yuan2006model} is a multi-task linear regression model that uses an $l_1/l_q$ norm as a regularization term to ensure common sparsity patterns among the regression weights of different tasks. %In other words, by training multiple linear regression models for each task it recognizes what set of features are relevant for a majority of them. 
In our experiments, we use the performance data for all the tasks in the XTREME-R \cite{ruder-etal-2021-xtreme} benchmark to train group lasso models.

\subsection{Results}

The average LOLO errors for the four tasks and the four methods are given in Table \ref{tab:main_results}. As we can see both Translated baseline and Performance Predictors can obtain much lower errors compared to the Average Score Baseline on PAWS-X, XNLI and XQUAD tasks. Group Lasso outperforms all the other methods on PAWS-X dataset while for XNLI and XQUAD datasets though, the Translate method outperforms the two performance predictor models. 

On TyDiQA-GoldP dataset , which had its test sets for different languages created independently without any translation, we see that the performance of Translate method drops with errors close to those obtained using the Average Score Baseline. While this behaviour is expected since the translated test sets and actual test sets now differ from each other, it still puts the reliability of the performance on translated data compared to the real data into question. Both XGBoost and Group Lasso though, obtain consistent improvements over the Baseline for TyDiQA-GoldP as well. 

\begin{figure}
    \centering
    \includegraphics[width=0.4\textwidth]{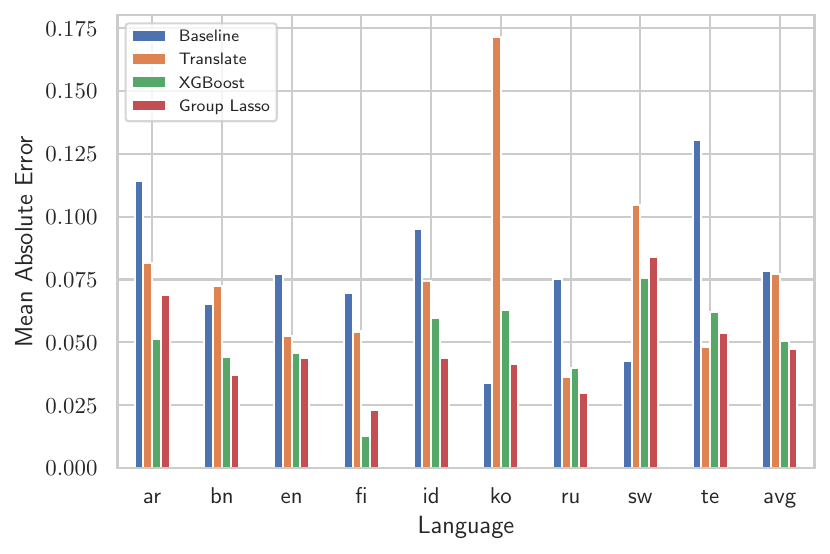}
    \caption{Language Wise Errors (LOLO setting) for predicting performances on the TyDiQA-GoldP dataset.}
    \label{fig:lang_wise_results}
\end{figure}

Figure \ref{fig:lang_wise_results} provides a breakdown of the errors for each language included in TyDiQA-GoldP benchmark, and again we can see that the Performance Predictors can outperform the Translate method almost all the languages except Telugu (te). Similar plots for the other tasks can be found in Figure \ref{fig:lang_wise_all} of Appendix. 

%Additionally, it should also be noted that translating test sets into different languages have costs attached to it as well as requires GPU compute time for running the evaluations. On the other hand, the performance predictors can be used to estimate the performance on all the 100 languages instantly at virtually no cost. 

%However, it must be noted that the original test sets of these tasks were also obtained by (manually) translating the English test sets and hence the pseudo test sets (obtained via automatic translation) are actually quite close to the original for most languages in these datasets. 

%On the other hand when we compare the errors on the TyDiQA-GoldP dataset, which had its test sets for different languages created independently without any translation, we see the performance of Translate method drops with errors close to those obtained using the Average Score Baseline. Both XGBoost and Group Lasso though, obtain consistent improvements over the Baseline for TyDiQA-GoldP as well, indicating that such approaches can overall provide a much more reliable estimation of the performance. 
% Figure \ref{fig:lang_wise_results} provides a breakdown of the errors for each language included in TyDiQA-GoldP benchmark, and again we can see that the Performance Predictors can outperform the Translate method almost all the languages except Telugu (te). Similar plots for the other 3 tasks can be found in Figure \ref{fig:lang_wise_all} of Appendix.

\subsection{Pivot Selection}
Another benefit of using Performance Prediction models is that we can use them to select training configurations like training (pivot) languages or amount of training data to achieve desired performance. For our case study we demonstrate the application of our predictors towards selecting the best pivot language for each of the 100 languages supported by mBERT that maximizes the predicted performance on the language. The optimization problem can be defined as:
\begin{equation}
    p^*(l) = \argmax_{p \in \mathcal{P}}f_{\theta}([\phi(l); \phi(p); \phi(p, l); \phi(\Delta_{tr}^{\{p\}})])
    \label{eq: opti}
\end{equation}

Where $p^*(l)$ denotes the pivot language that results in the best predicted performance on language $l \in \mathcal{L}$. Since, $\mathcal{P} = \{en\}$ only for PAWS-X, XQUAD and XNLI i.e. training data is available only in English, we run this experiment on TyDiQA-GoldP dataset which has training data available in 9 languages i.e. $\mathcal{P} = \{ar, bn, es, fi, id, ko, ru, sw, te\}$. We solve the optimization problem exactly by evaluating Equation \ref{eq: opti} for all $(p, l)$ pairs using a linear search and we use XGBoost Regressor as $f_{\theta}$.% $ \phi(\Delta_{tr})$ for a selected pivot language $p$ is kept fixed to the entire training set, and use XGBoost Regressor as $f_{\theta}$.

\begin{figure}
    \centering
        \includegraphics[width=0.4\textwidth]{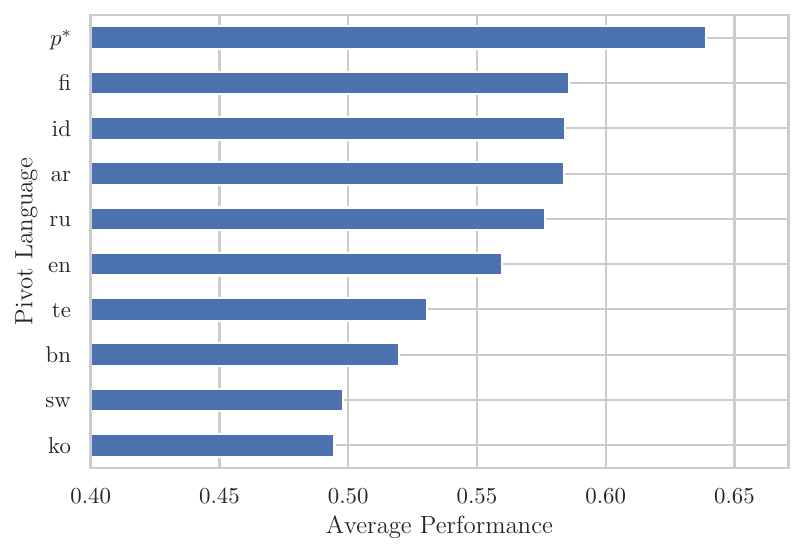}
        \caption{Average Performance on the 100 languages supported by mBERT for each of the 9 pivot languages for which training data is available in TyDiQA-GoldP.}
        \label{fig:pivot_avg_perf}
    % \begin{subfigure}[t][width=0.45\textwidth]
    %     \includegraphics[width=0.9\textwidth]{figures/pivot_wise_avg_performance.pdf}
    %     \caption{Average Performance on the 100 languages supported by mBERT for each of the 9 pivot languages for which training data is available in TyDiQA-GoldP.}
    %     \label{fig:pivot_avg_perf}
    % \end{subfigure}
    %\caption{Summary of results from Pivot Selection for TyDiQA-GoldP dataset.}
    %\label{fig:pivot_select}
\end{figure}

The results of this exercise are summarized in Figure \ref{fig:pivot_avg_perf}. We see carefully selecting the best pivot for each language leads to substantially higher estimated performances instead of using the same language as pivot for all the languages. We also see that languages like Finnish, Indonesian, Arabic and Russian have higher average predicted performance across all the supported languages compared to English. This observation is also in line with \citet{turc2021revisiting} observation that English might not always be the best pivot language for zero-shot transfer.

%The predictor models can also be used to do similar but more sophisticated search procedures like searching for the optimum data allocation strategy for multiple pivot languages to maximize the estimated performance under a fixed data collection budget as demonstrated by \citet{srinivasan2021predicting}. \citet{xia-etal-2020-predicting} also use a search procedure over the performance predictors to select the most representative set of languages that minimizes the error on a much larger set of languages.

% These results put into question the reliability of (manual or automatic) translated test sets for measuring the performance on a language for a given task. \citet{clark-2021-strong}

% 5. Guidelines: (Thinking out loud) Can we propose some guidelines around how to choose a set of languages to choose while creating a multilingual benchmark. Even though the predictor can help extrapolating to the other languages the best course of operation as has been observed in the previous work has been when the features at the test time do not significantly deviate from the features observed at the training time. Xia et al. uses a beam search procedure to select the best set of languages that gives the least RMSE on other languages. However this does assume the presence of all the set of languages in advance which might not be practical.

\section{Conclusion}

In this paper we discussed how the current state of benchmarking multilingual models is fundamentally limited by the amount of languages supported by the existing benchmarks, and proposed Performance Prediction as a potential solution to address the problem. Based on the discussion we summarize our findings through three key takeaways

\noindent \textbf{1.} Training performance prediction models on the existing evaluation data available for a benchmark can be a simple yet effective solution in estimating the MMLM's performance on a larger set of supported languages, which can often lead to much closer estimates compared to using the expected value estimate obtained from the existing languages.

\noindent \textbf{2.} One should be careful in using translated data to evaluate a model's performance on a language. Our experiments suggest that the performance measures estimated from the translated data can miscalculate the actual performance on the real world data for a language.

\noindent \textbf{3.} Performance Prediction can not only be effective for benchmarking on a larger set of languages but can also aid in selecting training strategies to maximize the performance of the MMLM on a given language which can be valuable towards building more accurate multilingual models.

Finally, there are a number of ways in which the current performance prediction methods can be improved for a more reliable estimation. Both \citet{xia-etal-2020-predicting,srinivasan2021predicting} observed that these models can struggle to generalize on languages or configurations that have features that are remarkably different from the training data. Multi-task learning as hinted by \citet{lin-etal-2019-choosing} and our experiments with Group Lasso can be a possible way to address this issue. The current methods also do not make use of model specific features for estimating the performance. \citet{tran2019transferability, nguyen2020leep, you2021logme} explore certain measures like entropy values, maximum evidence derived from a pre-trained model to estimate the transferability of the learned representations. It can be worth exploring if such measures can be helpful in providing more accurate predictions.

\bibliography{anthology,custom}
\bibliographystyle{acl_natbib}

\appendix
\clearpage
\section{Appendix}
\label{sec:appendix}
\begin{figure*}
    \centering
    \begin{subfigure}[t]{0.45\textwidth}
    \includegraphics[width=0.95\textwidth]{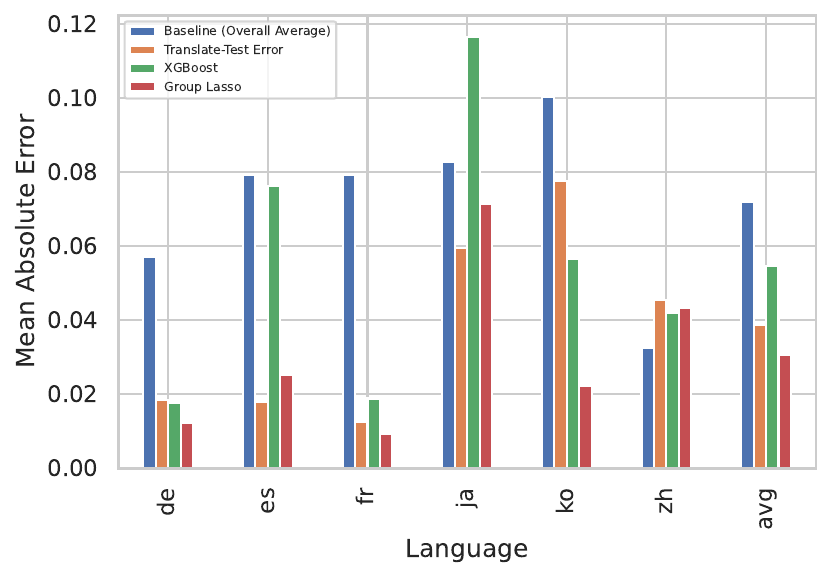}
    \caption{Language-Wise Errors for PAWS-X dataset.}
    \label{fig:lang_wise_paws}
    \end{subfigure}
    \begin{subfigure}[t]{0.45\textwidth}
    \includegraphics[width=0.95\textwidth]{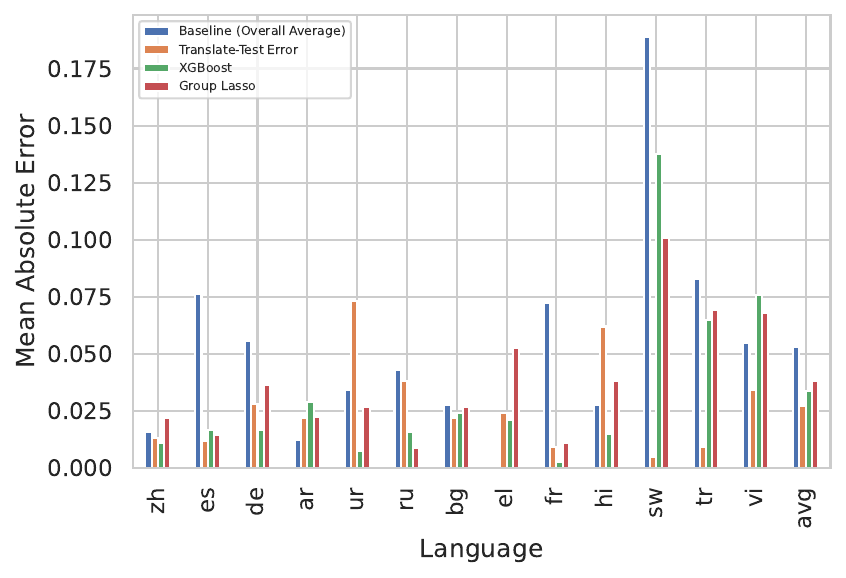}
    \caption{Language-Wise Errors for XNLI dataset.}
    \label{fig:lang_wise_xnli}
    \end{subfigure}
    \begin{subfigure}[t]{0.45\textwidth}
    \includegraphics[width=0.95\textwidth]{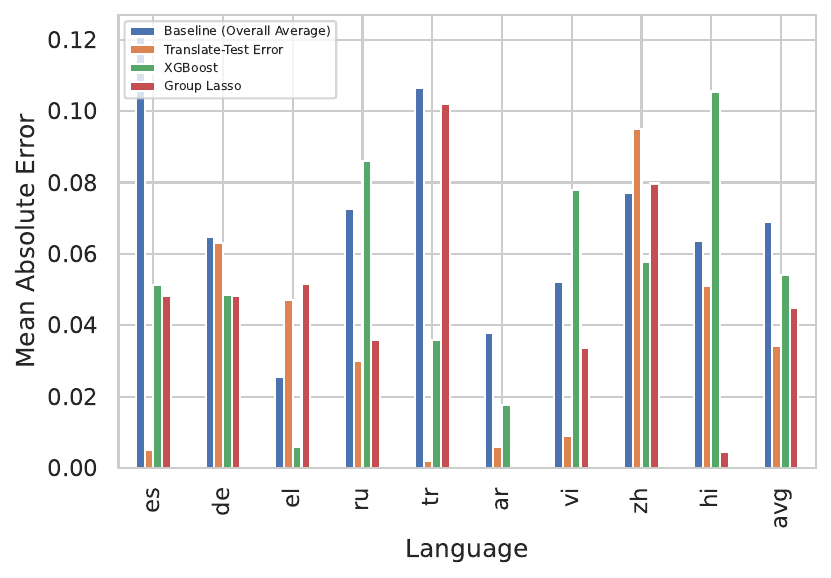}
    \caption{Language-Wise Errors for XQUAD dataset.}
    \label{fig:lang_wise_xquad}
    \end{subfigure}
\caption{}
\label{fig:lang_wise_all}
\end{figure*}

Table \ref{tab:all_tasks} contains the information about the tasks considered in the survey for Section \ref{sec:problem}. The language-wise errors for tasks other than TyDiQA-GoldP can be found in Figure \ref{fig:lang_wise_all}.

\subsection{Training Details}
\noindent
\textbf{Performance Prediction Models}
\begin{enumerate}
    \item {XGBoost}: For training XGBoost regressor for the performance prediction, we use 100 estimators with a maximum depth of 10 and a learning rate of 0.1.
    \item {Group Lasso}: We use a regularization strength of $0.005$ for the $l_1/l_2$ norm term in the objective function, and use the implementation provided in the MuTaR software package \footnote{https://github.com/hichamjanati/mutar}.
\end{enumerate}

\noindent
\textbf{Translate Baseline}:
We use the Azure Translator\footnote{https://azure.microsoft.com/en-us/services/cognitive-
services/translator/} to translate the data in pivot language to target languages. For classification tasks XNLI and PAWS-X, the labels can be directly transferred across the translations. For QA tasks XQUAD and TyDiQA we use the approach described in \citet{hu2020xtreme} to obtain the answer span in the translated test which involves enclosing the answer span in the original text within <b> </b> tags to recover the answer in the translation.

\subsection{Features Description}
\label{sec:feat_desc}
\noindent \textbf{1. Pre-training Size of a Language}: The amount of data in a language $l$ that was used to pre-train the MMLM.

\noindent \textbf{2. Tokenizer Quality}: We use the two metrics defined by \citet{rust-etal-2021-good} to measure the quality of a multilingual tokenizer on a target language $t$. The first metric is \textbf{Fertility} which is equal to the average number of sub-words produced per tokenized word and the other is \textbf{Percentage Continued Words} which measures
how often the tokenizer chooses to continue a
word across at least two tokens.

\noindent \textbf{3. Subword Overlap}: The subword overlap between a pivot and target language is defined as the fraction of sub-words that are common in the vocabulary of the two languages.  Let $V_p$ and $V_t$ be the subword vocabularies of $p$ and $t$. The subword overlap is then defined as :

\begin{equation}
    o_{sw}(p, t) = \frac{|V_p \cap V_t|}{|V_p    \cup V_t|}
\end{equation}

\noindent \textbf{4. Relatedness between Lang2Vec features}:
Following \citet{lin-etal-2019-choosing} and \citet{lauscher-etal-2020-zero}, we compute the typological relatedness between $p$ and $t$ from the linguistic features provided by the URIEL project \cite{littell-etal-2017-uriel}. 
We use syntactic ($s_{syn}(p,t)$), phonological similarity ($s_{pho}(p,t)$), genetic similarity ($s_{gen}(p,t)$) and geographic distance ($d_{geo}(p,t)$). For details, please see \citet{littell-etal-2017-uriel}

\begin{table*}
\centering
\begin{tabular}{p{3cm}p{3.5cm}p{2cm}p{2cm}p{2cm}}%{lllrr}
\toprule
{} &                     Type & Release Year &  Number of Languages &  Number of Language Families \\
\midrule
UDPOS \cite{nivre2018universal}      &     Structure Prediction &         2015 &         57 &                 13 \\
WikiANN \cite{pan-etal-2017-cross}    &     Structure Prediction &         2017 &        100 &                 15 \\
XNLI \cite{conneau-etal-2018-xnli}       &           Classification &         2018 &         15 &                  7 \\
XCOPA \cite{ponti-etal-2020-xcopa}      &           Classification &         2020 &         10 &                 10 \\
XQUAD \cite{artetxe-etal-2020-cross}    &       Question Answering &         2020 &         11 &                  6 \\
MLQA \cite{lewis-etal-2020-mlqa}     &       Question Answering &         2020 &          7 &                  4 \\
TyDiQA \cite{clark-etal-2020-tydi}     &       Question Answering &         2020 &         11 &                  9 \\
MewsliX \cite{ruder-etal-2021-xtreme}    &                Retrieval &         2020 &         11 &                  5 \\
LAReQA \cite{roy-etal-2020-lareqa}     &                Retrieval &         2020 &         11 &                  6 \\
PAWSX \cite{yang-etal-2019-paws}     &  Sentence Classification &         2019 &          7 &                  4 \\
BUCC \cite{zweigenbaum-etal-2017-overview}       &                Retrieval &         2016 &          4 &                  2 \\
MLDoc \cite{schwenk-li-2018-corpus}      &           Classification &         2018 &          8 &                  3 \\
QALD-9 \cite{perevalov-etal-2022-qald}     &       Question Answering &         2022 &          9 &                  2 \\
xSID \cite{van-der-goot-etal-2021-masked}       &           Classification &         2021 &         11 &                  6 \\
WikiNEuRal \cite{tedeschi-etal-2021-wikineural-combined} &     Structure Prediction &         2021 &          8 &                  1 \\
WikiLingua \cite{ladhak-etal-2020-wikilingua} &            Summarization &         2020 &         18 &                  9 \\
XL-BEL \cite{liu-etal-2021-learning-domain}     &                Retrieval &         2021 &         10 &                  7 \\
Tatoeba \cite{artetxe-schwenk-2019-massively}    &                Retrieval &         2019 &         73 &                 14 \\
\bottomrule
\end{tabular}
\caption{The list of tasks surveyed for the discussion in Section \ref{sec:problem}.}
\label{tab:all_tasks}
\end{table*}

\end{document}